\documentclass[conference]{IEEEtran}
\IEEEoverridecommandlockouts

\usepackage{cite}
\usepackage{amsmath,amssymb,amsfonts}
\usepackage{algorithmic}
\usepackage{graphicx}
\usepackage{textcomp}
\usepackage{xcolor}
\usepackage{booktabs}
\usepackage{xcolor}
\usepackage{colortbl}
\usepackage{amssymb} 
\usepackage{amsmath} 
\usepackage{bm}
\usepackage{multirow}
\usepackage{graphicx}
\usepackage[table]{xcolor}

\newcolumntype{C}{>{\centering\arraybackslash}X}
\usepackage{tabularx}
\usepackage{booktabs}
\usepackage{multirow}
\usepackage{makecell}
\usepackage[table]{xcolor}
\usepackage{graphicx}
\usepackage{makecell}
\usepackage{pdfrender}
\usepackage{enumitem}

\definecolor{yeoVis}{HTML}{781286}        
\definecolor{yeoSomMot}{HTML}{4682B4}     
\definecolor{yeoDorsAttn}{HTML}{00760E}   
\definecolor{yeoSalVentAttn}{HTML}{C43AFA}
\definecolor{yeoLimbic}{HTML}{DCCA4F}     
\definecolor{yeoCont}{HTML}{E69422}       
\definecolor{yeoDefault}{HTML}{CD3E4E}

\usepackage[T1]{fontenc}

\begin{document}

\title{CoHyFuse: Condition-wise Hypergraph Fusion with Global Connectome in Task-fMRI}

\author{
Boseong Kim\textsuperscript{1},
Haejun Chung\textsuperscript{1},
and Ikbeom Jang\textsuperscript{2}\\[0.5em]
\textsuperscript{1}Hanyang University, Seoul, Republic of Korea\\
\textsuperscript{2}Hankuk University of Foreign Studies, Yongin, Republic of Korea\\
Correspondence: ijang@hufs.ac.kr; haejun@hanyang.ac.kr
}

\maketitle

\begin{abstract}
Task-fMRI connectomes reveal state-dependent neural reconfigurations, yet conventional methods marginalize these signals by aggregating distinct conditions into static pairwise graphs, thereby obscuring condition-specific multi-ROI organization. We introduce CoHyFuse, a condition-aware ROI-centered hypergraph framework that constructs a task-state-specific incidence matrix from condition-wise functional connectivity (FC)-profile embeddings, allowing the same ROI to form different multi-ROI hyperedges across task phases. Condition-specific neighborhood sizes $K_q$ further adapt the hyperedge scale to each task state, and the resulting condition embeddings are fused with a complementary whole-session FC branch for prediction. In the AABC cohort (N=1,074), CoHyFuse achieved the best mean out-of-fold predictive performance among evaluated baselines on FACENAME Fluid Cognition Composite (FCC) prediction (7.83$\pm$0.10 MAE, 0.439$\pm$0.026 \(R^2\)) and VISMOTOR age prediction (7.52$\pm$0.37 MAE, 0.592$\pm$0.022 \(R^2\)). In an auxiliary CMI-HBN attention-deficit/hyperactivity disorder (ADHD) classification benchmark (N=223), CoHyFuse obtained 72.0$\pm$2.1\% macro-AUC and 74.2$\pm$2.9\% accuracy. Ablation studies support the contributions of condition-wise incidence construction and dual-view fusion, suggesting that state-resolved ROI-set structure provides complementary predictive information beyond whole-session FC alone. Occlusion analysis identifies the Distraction condition as the primary driver of model prediction, pointing toward the Salience/Ventral Attention Network (SAN)--FrontoParietal Network (FPN) and within-SAN hyperedge-defined ROI-set motifs as candidate model-relevant patterns. This framework provides an interpretable, state-resolved view of the connectome for downstream cohort analysis. Code will be released upon acceptance.
\end{abstract}
\vspace{-4pt}

\vspace{-4pt}

\section{Introduction}
\vspace{-4pt}
Population-scale neuroimaging initiatives provide task-based functional magnetic resonance imaging (task-fMRI) together with standardized behavioral and cognitive phenotyping, enabling connectome-based modeling of inter-individual variability. Accumulating evidence suggests that task-evoked brain states can increase behaviorally relevant variance in functional connectivity (FC) and, for some traits and paradigms, task-derived connectivity can improve out-of-sample behavioral prediction relative to resting-state connectivity \cite{greene2018task}. These findings motivate task-fMRI connectome models for cognitive phenotype prediction that preserve structured condition dynamics across paradigms and cohorts \cite{akshoomoff2013viii}.

Most connectome-learning approaches represent FC as pairwise graphs or FC matrices and learn a single, condition-agnostic representation of an entire scan \cite{li2021braingnn, wang2023adaptive}. However, task paradigms are inherently heterogeneous: FC reconfigures across conditions, and collapsing repeated phases into a single representation can attenuate condition-wise signatures. Moreover, pairwise graphs restrict message passing to dyadic region of interest (ROI)–ROI edges and do not directly aggregate information over ROI sets. Hypergraphs generalize graphs by allowing a hyperedge to connect multiple nodes; in this work, each hyperedge is operationally defined as an FC-profile neighborhood, enabling incidence-based aggregation over condition-specific ROI sets \cite{feng2019hypergraph}. Prior hypergraph connectomics studies suggest that such ROI-set topology can carry behaviorally and clinically relevant information \cite{yuan2024mhsa}. To address this gap, we propose CoHyFuse, a condition-aware ROI-centered hypergraph fusion framework that builds condition-wise top-$K_q$ hypergraphs. CoHyFuse integrates attention-fused hypergraph neural networks (HGNN) embeddings with a whole-session FC branch, capturing both condition-wise reconfiguration cues and session-level connectivity structure for task-fMRI prediction.

Our contributions are threefold:
\begin{enumerate}[label=(\roman*), leftmargin=*, itemsep=0.2em]
    \item We propose \textbf{CoHyFuse}, a condition-aware ROI-centered hypergraph fusion framework that constructs condition-wise top-$K_q$ hypergraphs and integrates attention-fused HGNN embeddings with a complementary whole-session FC branch, thereby capturing both condition-wise reconfiguration cues and session-level connectivity structure for task-fMRI prediction.

    \item We evaluate CoHyFuse on two public fMRI datasets, and CoHyFuse achieves the best mean performance among evaluated baselines. Ablation results support the effectiveness of condition-specific hypergraph construction, heterogeneous \(K_q\), and dual-branch fusion.
    

    \item Applying CoHyFuse to FACENAME, we find that the Distraction condition yields the strongest association between hyperedge-guided subnetwork strength and the FCC ($r{=}0.36$), with dominant network-pair contributions in SAN--SAN and SAN--FPN connectivity.

\end{enumerate}

\begin{figure*}[t]
    \centering
    \includegraphics[page=1, width=0.95\textwidth]{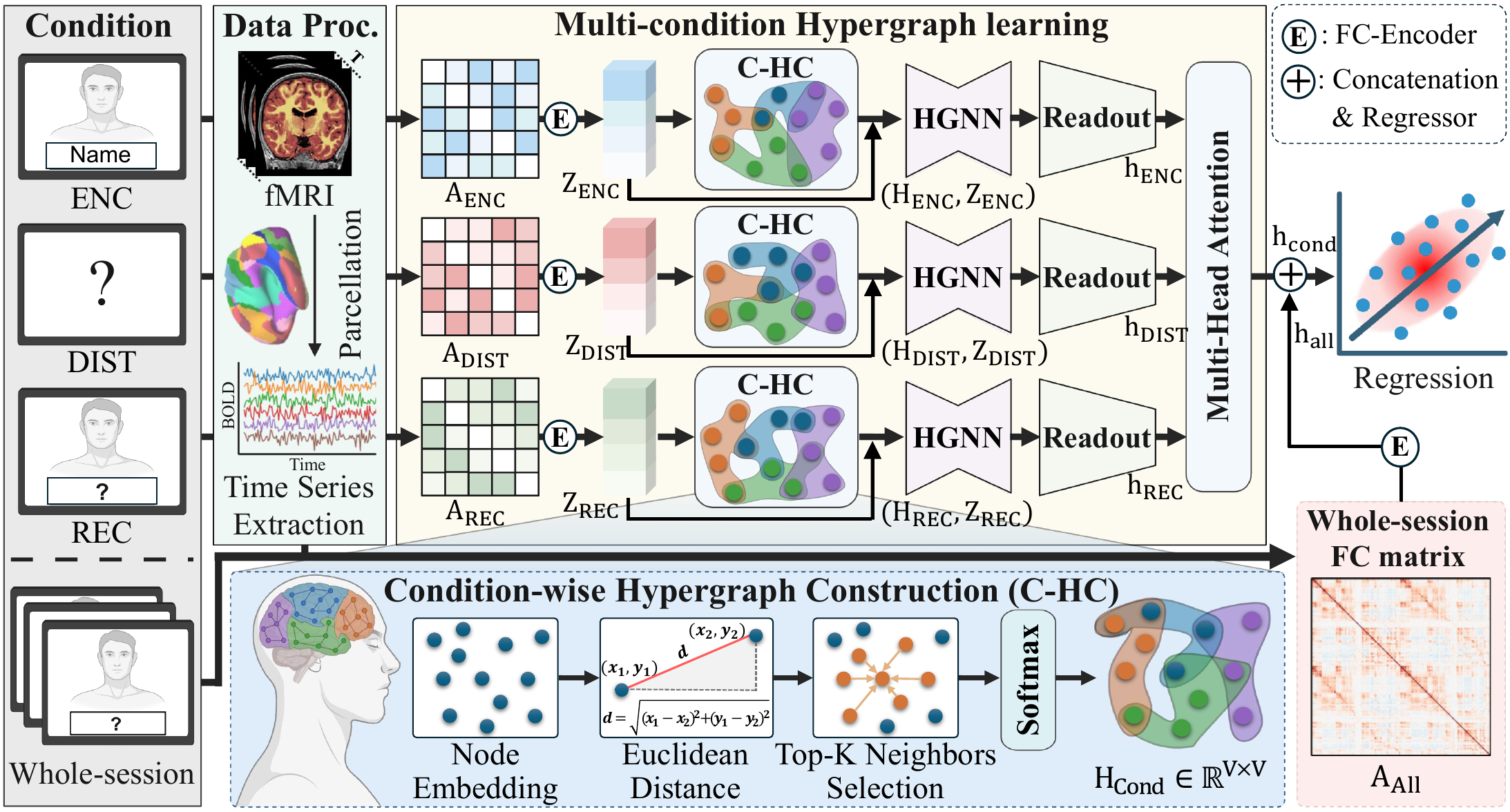}
    \vspace{-2mm}
    \caption{Overview of the proposed CoHyFuse. Condition-wise ROI-centered hypergraphs are encoded by HGNN branches, fused by multi-head attention and jointly integrated with a whole-session FC branch for prediction.}
    \label{fig:1}
    \vspace{-4mm}
\end{figure*}

\vspace{-3pt}
\section{Related Work}

\subsection{State-resolved connectivity in task-fMRI}
Task-fMRI provides structured brain states that can reveal behaviorally relevant individual differences in functional connectivity. Prior studies have shown that task-evoked connectivity can improve individual trait or behavioral prediction compared with resting-state connectivity \cite{greene2018task}. These findings establish the importance of preserving task context, but they mainly estimate seed-based or pairwise ROI–ROI connectivity rather than learning condition-specific ROI-set representations for prediction.


\subsection{Graph-based connectome learning}
Recent deep connectome models represent fMRI connectivity as graphs or FC matrices and learn predictive representations using graph neural networks (GNNs) or Transformers. GCN- and GAT-based models perform message passing over pairwise ROI--ROI edges, while brain-specific architectures such as BrainGNN introduce interpretable graph learning for fMRI analysis \cite{kipf2016semi,velivckovic2017graph,li2021braingnn}. Transformer-based connectome models further aim to capture long-range dependencies in FC matrices \cite{yu2024long}. Although these approaches improve representation learning over conventional machine learning baselines, most of them still operate on whole-session or condition-agnostic pairwise connectivity. As a result, they may miss task-state-specific reconfiguration patterns and condition-specific multi-ROI organization.

\subsection{Hypergraph learning for ROI-set connectomics}
Hypergraphs extend conventional graphs by allowing one hyperedge to connect multiple ROIs, providing a natural representation for ROI-set structure \cite{feng2019hypergraph}. Hypergraph neural networks perform incidence-based message passing through node–hyperedge memberships \cite{feng2019hypergraph}, and later hypergraph models further improve hyperedge-aware propagation and representation learning \cite{dong2020hnhn,wang2022equivariant}. Recent hypergraph connectomics studies suggest that multi-ROI topological or representational patterns can improve task decoding, individual identification, and brain–behavior association \cite{yuan2024mhsa}. However, existing hypergraph connectome approaches generally do not explicitly construct task-condition-specific hypergraphs or adapt the neighborhood scale across task states. The key distinction is that task condition is not treated merely as an additional feature or branch label; it determines the incidence structure itself, so that ROI-set membership is allowed to change across paradigm-defined brain states.

CoHyFuse addresses this gap by building condition-wise ROI-centered top-$K_q$ hypergraphs and fusing their HGNN embeddings with a complementary whole-session FC branch, thereby linking condition-resolved task-fMRI modeling with hypergraph-based ROI-set connectome learning.

\section{Methods}

\subsection{Overview and Problem Formulation}
Given an ROI time series $\mathbf{X}\in\mathbb{R}^{T\times V}$ and task-condition labels
$\mathbf{c}=\{c_t\}_{t=1}^{T}$ where $c_t\in\{1,\dots,Q\}$ denotes the condition at repetition time (TR) $t$,
we learn a regressor $f_{\Theta}$ to predict $y\in\mathbb{R}$:
\begin{equation}
\hat{y}=f_{\Theta}(\mathbf{X},\mathbf{c}).
\end{equation}
Here, c denotes paradigm timing information available from the task-fMRI design, not subject-level metadata or target-derived information; CoHyFuse uses it only to partition time points into task states and to construct state-resolved FC/hypergraph representations. Our framework first estimates condition-wise FC and constructs a top-$K_q$ neighborhood hypergraph for each condition. An HGNN encodes each condition-wise hypergraph to obtain condition embeddings, which are integrated via multi-head self-attention. Finally, the fused condition representation is combined with a whole-session FC feature vector to predict $\hat{y}$ (Fig.~\ref{fig:1}).

\subsection{Condition-wise Hypergraph Construction}
Let $\mathbf{X}\in\mathbb{R}^{T\times V}$ be the ROI time series and $\mathbf{c}=\{c_t\}_{t=1}^{T}$ the paradigm-defined task-condition labels, where $c_t\in\{1,\dots,Q\}$. For each condition $q$, we define $\mathcal{T}_q=\{t~|~c_t=q\}$ and estimate a condition-wise FC matrix $\mathbf{A}^{(q)}\in\mathbb{R}^{V\times V}$ by computing Pearson correlations between ROI time series over the time points in $\mathcal{T}_q$. We set the diagonal to zero and apply Fisher $z$-transform to off-diagonal correlations. The samples in \(T_q\) are treated as condition-indexed observations pooled across repeated blocks, rather than as a newly continuous time series across block boundaries. The resulting condition-wise FCs are used as FC-profile features for ROI-set neighborhood construction and prediction; interpretation is therefore focused on hyperedge/subnetwork-level patterns rather than individual FC edges.

Each ROI is treated as a node. For condition \(q\), node \(i\) is initialized with its FC profile \(\mathbf r_i^{(q)}=\mathbf A_{i,:}^{(q)}\) and embedded by a trainable FC-profile encoder as \(\mathbf z_i^{(q)}=\phi_\theta(\mathbf r_i^{(q)})\). We define each hyperedge as an FC-profile neighborhood and use the resulting node--hyperedge incidence structure for message passing. This construction yields representation-level ROI-set neighborhoods based on shared condition-wise connectivity profiles. We instantiate $E=V$ hyperedges per condition, one anchored at each ROI (column $i$ is anchored at ROI $i$).
We compute pairwise similarity in the embedding space using negative squared Euclidean distance:
\vspace{-3pt}
\begin{equation}
s^{(q)}_{ij} = -\|\mathbf{z}^{(q)}_i-\mathbf{z}^{(q)}_j\|_2^2.
\end{equation}
\vspace{-3pt}
For each anchor node $i$, we select the top-$K_q$ neighbors $\mathcal{N}_{K_q}(i)$ among $j \ne i$ and define a weighted incidence matrix
$\mathbf{H}^{(q)}\in\mathbb{R}^{V\times V}$ as
\vspace{-3pt}
\begin{equation}
H^{(q)}_{j,i}=
\begin{cases}
1, & j=i,\\
K_q\cdot \left[\mathrm{softmax}_{j'\in\mathcal{N}_{K_q}(i)}\!\left(\dfrac{s^{(q)}_{ij'}}{\tau_q}\right)\right]_{j}, 
& j\in\mathcal{N}_{K_q}(i),\\
0, & \text{otherwise},
\end{cases}
\end{equation}


where \(\tau_q>0\) is a learnable condition-wise temperature. Although the support is obtained by a top-$K_q$ search, the selected set is not used as independent dyadic ROI--ROI edges. Each selected neighborhood defines one column of the incidence matrix $H^{(q)}$, which is then used by the HGNN layer for node--hyperedge--node propagation in Sec.~\ref{sec:multicond}. Thus, the top-$K_q$ step specifies condition-specific ROI-set memberships, whereas the predictive operator retains an incidence-based aggregation structure.

In our implementation, \(\mathbf H^{(q)}\) is recomputed from the current embeddings at each forward pass; the top-\(K_q\) indices define a non-differentiable support for that pass, so gradients are not propagated through the index-selection step but flow through the selected softmax weights and downstream modules. We use condition-wise neighborhood sizes $K_q$ rather than a single shared $K$. We multiply the softmax neighbor memberships by $K_q$ so that the total neighbor mass scales with $K_q$, preventing trivial shrinkage of per-neighbor weights when larger neighborhoods are used. 


\subsection{Multi-condition Hypergraph Learning}
\label{sec:multicond}
Given $\{(\mathbf{H}^{(q)},\mathbf{Z}^{(q)})\}_{q=1}^{Q}$, we encode each condition-wise hypergraph and integrate the resulting condition embeddings to predict $y$.

\subsubsection{Condition-wise HGNN}
For each condition $q$, we apply a hypergraph convolution with symmetric degree normalization \cite{feng2019hypergraph}. We use condition-wise branches with shared HGNN parameters $\mathbf{\Theta}$ across conditions:
\begin{equation}
\mathbf{X}^{(q)}
=\sigma\!\left(
\mathbf{D}^{(q)}_{v}{}^{-\frac{1}{2}}
\mathbf{H}^{(q)}
\mathbf{D}^{(q)}_{e}{}^{-{1}}
\mathbf{H}^{(q)\top}
\mathbf{D}^{(q)}_{v}{}^{-\frac{1}{2}}
\mathbf{Z}^{(q)}\mathbf{\Theta}
\right),
\end{equation}
where $\mathbf{D}^{(q)}_{v}=\mathrm{diag}(\mathbf{H}^{(q)}\mathbf{1})$ and
$\mathbf{D}^{(q)}_{e}=\mathrm{diag}(\mathbf{H}^{(q)\top}\mathbf{1})$.
We then obtain a condition embedding by attention pooling over ROIs:
\begin{equation}
\alpha^{(q)}=\mathrm{softmax}(\psi(\mathbf{X}^{(q)})), \qquad
\mathbf{h}^{(q)}=\sum_{i=1}^{V}\alpha^{(q)}_i\,\mathbf{X}^{(q)}_{i,:}\in\mathbb{R}^{D},
\end{equation}
where $\psi(\cdot)$ is an MLP producing a scalar attention logit per ROI.

\subsubsection{Multi-condition fusion}
We stack condition embeddings as $\mathbf{E}_c=[\mathbf{h}^{(1)},\dots,$ $\mathbf{h}^{(Q)}]^{\top}\in\mathbb{R}^{Q\times D}$ and
apply a self-attention fusion block \cite{vaswani2017attention} (multi-head self-attention followed by a feed-forward network with residual connections),
yielding $\tilde{\mathbf{E}}_c\in\mathbb{R}^{Q\times D}$. We obtain a fused condition representation by mean pooling,
$\mathbf{h}_{\mathrm{cond}}=\frac{1}{Q}\sum_{q=1}^{Q}\tilde{\mathbf{E}}_c(q,:)\in\mathbb{R}^{D}$.
In parallel, we compute a whole-session FC matrix $\mathbf{A}^{(\mathrm{all})}$ from all TRs and use the vectorized upper-triangular
entries $\mathbf{a}=\mathrm{vec}(\mathbf{A}^{(\mathrm{all})}_{\mathrm{triu}})\in\mathbb{R}^{P}$, where $P=V(V-1)/2$,
encoded by an MLP to $\mathbf{h}_{\mathrm{all}}\in\mathbb{R}^{D}$.
The condition-wise hypergraphs capture context-dependent reconfiguration, while the whole-session FC captures stable session-level connectivity structure. We fuse both representations to leverage dynamic and stable cues for prediction. The final prediction is
\begin{equation}
\hat{y}=g\!\left([\mathbf{h}_{\mathrm{cond}};\mathbf{h}_{\mathrm{all}}]\right),
\end{equation}
where $g(\cdot)$ is an MLP regressor. The model is trained by minimizing MSE between $\hat{y}$ and $y$.

\begin{table}[t]
\caption{Demographics of the AABC cohort.}
\vspace{-1.5mm}
\centering
\renewcommand{\arraystretch}{0.90}
\setlength{\tabcolsep}{3.0pt}
\resizebox{0.85\linewidth}{!}{%
\begin{tabular}{@{}cccc@{}}
\toprule
\textbf{\shortstack{Subjects (N)}} 
& \textbf{\shortstack{Sex (M/F)}} 
& \textbf{\shortstack{Age (Mean$\pm$Std)}} 
& \textbf{\shortstack{FCC (Mean$\pm$Std)}} \\
\midrule
1,074 
& 495 / 579 
& $60.53\pm15.16$ 
& $96.41\pm13.34$ \\
\bottomrule
\end{tabular}%
}
\label{tab:aabc_demographics}
\vspace{-4mm}
\end{table}

\begin{table}[t]
\caption{Demographics of the CMI-HBN cohort.}
\vspace{-1.5mm}
\centering
\renewcommand{\arraystretch}{0.90}
\setlength{\tabcolsep}{4.5pt}
\resizebox{0.85\linewidth}{!}{%
\begin{tabular}{@{}lccc@{}}
\toprule
\textbf{Group} 
& \textbf{Subjects (N)}
& \textbf{Sex (M/F)}
& \textbf{Age (Mean$\pm$Std)} \\
\midrule
NoDx   
& 68  
& 37 / 31 
& $8.86\pm1.75$ \\

ADHD-I 
& 75 
& 45 / 30 
& $9.48\pm1.45$ \\

ADHD-C 
& 80 
& 42 / 38 
& $8.58\pm1.76$ \\
\bottomrule
\end{tabular}%
}
\label{tab:cmihbn_demographics}
\vspace{-5mm}
\end{table}

\section{Experiments and Results}

\subsection{Dataset and Experiment Settings}

\subsubsection{Dataset}

\begin{table*}[t]
\caption{Regression results (mean$\pm$std) on FACENAME and VISMOTOR task-fMRI. Best and second-best are \textbf{bold} and \underline{underlined}.}
\vspace{-1.5mm}
\centering
\renewcommand{\arraystretch}{0.95}
\setlength{\tabcolsep}{4pt}
\resizebox{0.98\textwidth}{!}{%
\begin{tabular}{l c ccc ccc}
\toprule
\multirow[c]{2}{*}{\textbf{Model}} 
& \multirow[c]{2}{*}{\textbf{Year}}
& \multicolumn{3}{c}{\textbf{FACENAME}}
& \multicolumn{3}{c}{\textbf{VISMOTOR}} \\
\cmidrule(lr){3-5} \cmidrule(lr){6-8}
&
& MAE
& RMSE
& $R^2$
& MAE
& RMSE
& $R^2$ \\
\midrule

SVM 
& 1995
& $8.21\!\pm\!0.15$ & $10.35\!\pm\!0.32$ & $0.386\!\pm\!0.036$
& $8.57\!\pm\!0.54$ & $10.83\!\pm\!0.64$ & $0.475\!\pm\!0.045$ \\

MLP
& 1986
& $8.31\!\pm\!0.10$ & $10.45\!\pm\!0.21$ & $0.374\!\pm\!0.036$
& $8.20\!\pm\!0.52$ & $10.09\!\pm\!0.65$ & $0.544\!\pm\!0.046$ \\

GCN \cite{kipf2016semi}
& 2017
& $8.32\!\pm\!0.11$ & $10.49\!\pm\!0.26$ & $0.369\!\pm\!0.033$
& $8.24\!\pm\!0.69$ & $10.44\!\pm\!0.95$ & $0.510\!\pm\!0.046$ \\

GAT \cite{velivckovic2017graph}
& 2018
& $8.29\!\pm\!0.15$ & $10.48\!\pm\!0.29$ & $0.370\!\pm\!0.032$
& $8.17\!\pm\!0.38$ & $10.10\!\pm\!0.32$ & $0.544\!\pm\!0.034$ \\

HGNN \cite{feng2019hypergraph}
& 2019
& $8.25\!\pm\!0.19$ & $10.46\!\pm\!0.38$ & $0.373\!\pm\!0.044$
& $8.05\!\pm\!0.38$ & $10.09\!\pm\!0.45$ & $0.544\!\pm\!0.038$ \\

HNHN \cite{dong2020hnhn}
& 2020
& $8.20\!\pm\!0.13$ & $10.34\!\pm\!0.33$ & $0.387\!\pm\!0.040$
& $8.01\!\pm\!0.62$ & $10.00\!\pm\!0.71$ & $0.552\!\pm\!0.040$ \\

BrainGNN \cite{li2021braingnn}
& 2021
& $8.43\!\pm\!0.35$ & $10.67\!\pm\!0.38$ & $0.346\!\pm\!0.039$
& $8.04\!\pm\!0.53$ & $9.99\!\pm\!0.67$ & $0.552\!\pm\!0.040$ \\

ED-HNN \cite{wang2022equivariant}
& 2022
& $8.38\!\pm\!0.27$ & $10.61\!\pm\!0.41$ & $0.355\!\pm\!0.046$
& $8.02\!\pm\!0.44$ & $9.95\!\pm\!0.48$ & $0.556\!\pm\!0.021$ \\

ALTER \cite{yu2024long}
& 2024
& $8.34\!\pm\!0.14$ & $10.48\!\pm\!0.21$ & $0.370\!\pm\!0.019$
& \underline{$7.91\!\pm\!0.36$} & \underline{$9.92\!\pm\!0.36$} & \underline{$0.559\!\pm\!0.026$} \\

TA-GAT \cite{wang2025towards}
& 2025
& \underline{$8.17\!\pm\!0.29$} & $10.40\!\pm\!0.42$ & $0.380\!\pm\!0.046$
& $8.13\!\pm\!0.45$ & $10.06\!\pm\!0.65$ & $0.546\!\pm\!0.038$ \\

STNAGNN \cite{wang2026stnagnn}
& 2026 
& $8.19\!\pm\!0.16$ & \underline{$10.30\!\pm\!0.33$} & \underline{$0.391\!\pm\!0.041$}
& $7.98\!\pm\!0.57$ & $10.01\!\pm\!0.54$ & $0.552\!\pm\!0.023$ \\

\rowcolor{gray!15}
\rule{0pt}{4.5ex}%
\textbf{\shortstack[l]{CoHyFuse\\(Proposed)}}
& \raisebox{1.2ex}{--}
& \shortstack{$\mathbf{7.83\!\pm\!0.10}$\\
\textcolor{blue}{$\mathbf{(0.34\downarrow)}$}}
& \shortstack{$\mathbf{9.91\!\pm\!0.25}$\\
\textcolor{blue}{$\mathbf{(0.39\downarrow)}$}}
& \shortstack{$\mathbf{0.439\!\pm\!0.026}$\\
\textcolor{blue}{$\mathbf{(0.048\uparrow)}$}}
& \shortstack{$\mathbf{7.52\!\pm\!0.37}$\\
\textcolor{blue}{$\mathbf{(0.39\downarrow)}$}}
& \shortstack{$\mathbf{9.61\!\pm\!0.36}$\\
\textcolor{blue}{$\mathbf{(0.31\downarrow)}$}}
& \shortstack{$\mathbf{0.592\!\pm\!0.022}$\\
\textcolor{blue}{$\mathbf{(0.033\uparrow)}$}} \\

\bottomrule
\end{tabular}%
}
\label{tab:benchmark_datasets}
\vspace{-4mm}
\end{table*}

We evaluated CoHyFuse using two publicly available fMRI datasets: Aging Adult Brain Connectome (AABC) Release 1 \cite{Bookheimer2019_HCPAging} and the Child Mind Institute Healthy Brain Network (CMI-HBN) \cite{alexander2017open}. Table~\ref{tab:aabc_demographics} summarizes the AABC analytic cohort used for both FACENAME and VISMOTOR benchmarks, and Table~\ref{tab:cmihbn_demographics} summarizes the CMI-HBN diagnostic cohort.

\noindent\textbf{AABC:} From the AABC cohort, we restricted analyses to the cross-sectional V1 visit and used two paradigms, FACENAME and VISMOTOR, as separate prediction benchmarks. After excluding participants with missing targets or unusable runs, we analyzed 1,074 adults aged 36--100+ years. Using the released HCP-style preprocessing \cite{glasser2013minimal}, we extracted ROI-mean time series with the Schaefer 100-region atlas \cite{schaefer2018local}. Targets were the NIH Toolbox Fluid Cognition Composite (FCC) for FACENAME \cite{akshoomoff2013viii} and chronological age for VISMOTOR. For condition-wise hypergraphs, timing files were converted to TR-wise condition labels, and valid same-condition samples were pooled across repeated blocks before FC estimation. FACENAME (345 volumes; TR$=$0.8\,s) has encoding (ENC), distraction (DIST), and recall (REC) (4$\times$25 TR blocks each; 100 TR/condition), excluding instruction/cue and run-in/out rest from FC estimation. VISMOTOR (194 volumes; TR$=$0.8\,s) has active/rest, where active blocks are 35/34/34 TRs (103 TR total) and intervening rest blocks define rest.

\noindent\textbf{CMI-HBN:} From the CMI-HBN pediatric cohort (ages 5--21) \cite{alexander2017open}, we restricted analyses to children aged 6--10 years from the complete-pass release with structural and blood-oxygen-level-dependent (BOLD) data passing quality control (QC), to reduce developmental heterogeneity. For functional QC, we required a median framewise displacement (FD) $\leq$ 0.2 mm and a normalized cross-correlation to template $\geq$ 0.8. After QC, age, sex, available site/scanner information, and median FD were inspected as cohort-level nuisance variables but were not used as model inputs; all learning-based models used identical subject-wise splits. Across the diagnostic groups, the Reproducible Brain Charts (RBC)-provided per-subject median FD did not differ significantly (Kruskal--Wallis, p=0.852; ADHD-I vs.\ ADHD-C, Mann--Whitney, p=0.562), suggesting that group-level motion differences were unlikely to drive classification. Participants were classified into three groups: no diagnosis (NoDx), ADHD inattentive type (ADHD-I), and ADHD combined type (ADHD-C). REST and PEER run labels were converted to TR-wise state labels, and valid same-state samples were pooled before FC estimation. PEER is an MR-based eye-calibration task with 27 fixation events (4\,s each; 5 TRs/event at TR$=$0.8\,s; 135 volumes/run); PEER1/PEER3 defined repeated PEER-state runs, and length-matched REST segments defined the REST state.

\subsubsection{Experimental Settings and Implementation Details}
All experiments were run on a single NVIDIA RTX A6000 GPU. We performed subject-wise 5-fold cross-validation with a 64\%/16\%/20\% train/validation/test split; all hyperparameter and checkpoint selections, including \(K_q\), used only the validation split, and the outer test fold was held out for final evaluation. CoHyFuse used a two-layer FC-profile encoder with output dimension \(D=64\), one shared-parameter HGNN layer per condition, 4-head self-attention for condition fusion, a two-layer whole-session FC MLP branch, and a two-layer prediction head, with ReLU activation, dropout \(p=0.2\), and approximately 0.71M trainable parameters. Models were optimized with AdamW using batch size 64, weight decay \(10^{-4}\), gradient clipping with max-norm \(=1.0\), and early stopping on validation loss for up to 80 epochs. The learning rate was selected from \(\{3\times10^{-4},5\times10^{-4},10^{-3}\}\), and \(K_q\) was searched over \(\{5,10,\ldots,50\}\) with step size 5. Regression used MSE loss, whereas the CMI-HBN classifier used cross-entropy loss. We report mean$\pm$std across folds.

\subsection{Comparison with Baseline Methods}
We compared CoHyFuse with classical baselines, including machine learning models (support vector machine (SVM) and multilayer perceptron (MLP)), graph neural networks (graph convolutional network (GCN) \cite{kipf2016semi}, graph attention network (GAT) \cite{velivckovic2017graph}, BrainGNN \cite{li2021braingnn}), recent task-fMRI-oriented GNN baselines, including TA-GAT \cite{wang2025towards} and STNAGNN \cite{wang2026stnagnn}, a Transformer-based method (ALTER \cite{yu2024long}), and hypergraph neural networks (HGNN \cite{feng2019hypergraph}, HNHN \cite{dong2020hnhn}, and ED-HNN \cite{wang2022equivariant}). All baselines were evaluated using the same subject-wise folds, preprocessing, ROI parcellation, and validation-based model selection. For TA-GAT and STNAGNN, available paradigm/state information was used following their original task-aware designs; CoHyFuse used the same timing only to form condition-wise FC profiles and task-state-specific incidence matrices.

As shown in Table~\ref{tab:benchmark_datasets}, CoHyFuse achieved the best mean performance among the evaluated baselines on both AABC tasks, reaching 7.83$\pm$0.10 MAE / 9.91$\pm$0.25 RMSE / 0.439$\pm$0.026 $R^2$ on FACENAME (improving MAE by 0.34 over TA-GAT and RMSE/$R^2$ by 0.39/0.048 over STNAGNN) and 7.52$\pm$0.37 MAE / 9.61$\pm$0.36 RMSE / 0.592$\pm$0.022 $R^2$ on VISMOTOR (improving by 0.39 MAE, 0.31 RMSE, and 0.033 $R^2$ over ALTER). Paired bootstrap analysis on out-of-fold predictions further supported the FACENAME MAE gain over TA-GAT ($\Delta$MAE = 0.34, 95\% CI [0.11, 0.57], $p=0.004$), and the VISMOTOR gain over ALTER was likewise significant ($\Delta$MAE = 0.39, 95\% CI [$0.18$, 0.60], $p=0.018$). Because FACENAME was evaluated in a broad adult lifespan cohort, we interpret the reported FACENAME $R^2$ as unadjusted total out-of-fold predictive performance rather than a covariate-adjusted effect size. A dedicated confound analysis in Sec.~\ref{sec:confound} showed that an age/sex/FD-only model was lower than CoHyFuse ($R^2=0.25$ vs. $0.439$) and that the FACENAME association remained significant after fold-wise confound adjustment (partial $r=0.34$, $p<0.001$). Notably, CoHyFuse also outperformed the other baselines on FACENAME, suggesting that explicit condition-wise hypergraph-based ROI-set fusion provides complementary predictive information beyond task-context-aware pairwise/spatiotemporal graph modeling.

\vspace{-4pt}
\subsection{Ablation Study}

\begin{table}[t]
\caption{Component ablation results on the FACENAME task.}
\vspace{-1.5mm}
\centering
\renewcommand{\arraystretch}{1}
\setlength{\tabcolsep}{3.8pt}
\resizebox{0.80\linewidth}{!}{%
\begin{tabular}{l c c}
\toprule
\textbf{Method} 
& MAE 
& $R^2$ \\
\midrule

w/o Whole FC 
& $8.20\!\pm\!0.26$ 
& $0.375\!\pm\!0.050$ \\

w/o Cond. branch
& $8.31\!\pm\!0.10$ 
& $0.374\!\pm\!0.036$ \\

w/o ENC cond. 
& $8.10\!\pm\!0.17$ 
& $0.397\!\pm\!0.036$ \\

w/o DIST cond. 
& $8.17\!\pm\!0.25$ 
& $0.397\!\pm\!0.031$ \\

w/o REC cond. 
& $8.08\!\pm\!0.15$ 
& $0.398\!\pm\!0.036$ \\

\rowcolor{gray!15}
\rule{0pt}{2ex}\textbf{Proposed} 
& $\mathbf{7.83\!\pm\!0.10}$ 
& $\mathbf{0.439\!\pm\!0.026}$ \\

\bottomrule
\end{tabular}%
}
\label{tab:component_ablation}
\vspace{-4mm}
\end{table}

\begin{table}[t]
\caption{Within-framework control results on the FACENAME task.}
\vspace{-1.5mm}
\centering
\renewcommand{\arraystretch}{1}
\setlength{\tabcolsep}{3.8pt}
\resizebox{0.80\linewidth}{!}{%
\begin{tabular}{l c c}
\toprule
\textbf{Variant} 
& MAE 
& $R^2$ \\
\midrule

CoHyFuse-MLP
& $8.11\!\pm\!0.19$
& $0.404\!\pm\!0.023$ \\

CoHyFuse-GCN 
& $8.13\!\pm\!0.14$ 
& $0.400\!\pm\!0.033$ \\

CoHyFuse-GAT 
& $8.07\!\pm\!0.13$ 
& $0.406\!\pm\!0.028$ \\

\rowcolor{gray!15}
\rule{0pt}{2ex}\textbf{CoHyFuse-HGNN} 
& $\mathbf{7.83\!\pm\!0.10}$ 
& $\mathbf{0.439\!\pm\!0.026}$ \\

\bottomrule
\end{tabular}%
}
\label{tab:operator_control}
\vspace{-4.5mm}
\end{table}

\subsubsection{Component-wise ablation of the proposed model}

Table~\ref{tab:component_ablation} presents the component ablation results, where one branch/component is removed at a time. Here, $K(E,D,R)$ denotes \((K_{\scriptscriptstyle\mathrm{ENC}},K_{\scriptscriptstyle\mathrm{DIST}},K_{\scriptscriptstyle\mathrm{REC}})\). Removing all condition branches (w/o Cond. branch) causes the largest drop ($\Delta$MAE $=+0.48$, $\Delta R^2=-0.065$). Removing the whole-session FC branch (w/o Whole FC) also degrades performance ($\Delta$MAE $=+0.37$, $\Delta R^2=-0.064$). Among single-condition removals, w/o DIST cond. yields the largest MAE increase ($+0.34$), followed by w/o ENC cond. ($+0.27$) and w/o REC cond. ($+0.25$).

\subsubsection{Effect of hypergraph aggregation}

Table~\ref{tab:operator_control} provides within-framework controls for separating condition-wise input effects from hypergraph aggregation. CoHyFuse-MLP uses vectorized condition-wise FC matrices and the whole-session FC matrix with an MLP predictor. Its performance shows that condition-wise FC inputs are informative, but CoHyFuse-HGNN further improves MAE from 8.11 to 7.83 and \(R^2\) from 0.404 to 0.439. CoHyFuse-GCN and CoHyFuse-GAT keep the same condition-wise inputs, $K_q$ search space, fusion structure, and whole-session FC branch, but replace the incidence-based HGNN operator with matched dyadic GCN/GAT propagation. These results support that the proposed condition-indexed incidence construction and hyperedge-based ROI-set aggregation provide additional predictive value beyond simple condition-wise FC concatenation or matched pairwise propagation.

\vspace{-5pt}
\subsection{Robustness Analysis}

\subsubsection{Effect of condition-wise $K_q$ in hypergraph construction}

\begin{table}[t]
\caption{Top-3 $K$ settings on the FACENAME task-fMRI.}
\vspace{-1.5mm}
\centering
\renewcommand{\arraystretch}{1}
\setlength{\tabcolsep}{3.8pt}
\resizebox{0.95\linewidth}{!}{%
\begin{tabular}{l c c c}
\toprule
\textbf{$K$ Scheme} 
& $K(E,D,R)$ 
& MAE 
& $R^2$ \\
\midrule

\multirow{3}{*}{Shared-$K$}
& $(5,\,5,\,5)$ & $7.98\!\pm\!0.09$ & $0.408\!\pm\!0.036$ \\
& $(10,\,10,\,10)$ & $8.03\!\pm\!0.14$ & $0.405\!\pm\!0.035$ \\
& $(50,\,50,\,50)$ & $8.05\!\pm\!0.12$ & $0.406\!\pm\!0.029$ \\

\midrule

\multirow{3}{*}{Cond-$K_q$}
& $(45,\,50,\,5)$ & $7.85\!\pm\!0.15$ & $0.433\!\pm\!0.024$ \\

& \cellcolor{gray!15}\rule{0pt}{2ex}$\mathbf{(50,\,45,\,5)}$ 
& \cellcolor{gray!15}$\mathbf{7.83\!\pm\!0.10}$ 
& \cellcolor{gray!15}$\mathbf{0.439\!\pm\!0.026}$ \\

& $(50,\,50,\,10)$ & $7.84\!\pm\!0.11$ & $0.432\!\pm\!0.025$ \\

\bottomrule
\end{tabular}%
}
\label{tab:facename_k_settings}
\vspace{-3mm}
\end{table}

\begin{figure}[t]
    \centering
    \includegraphics[width=1\columnwidth]{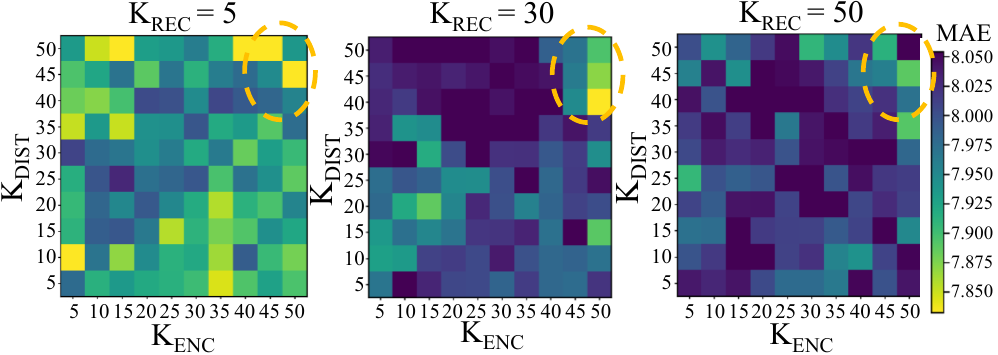}
    \vspace{-6mm}
    \caption{$K_q$ sensitivity heatmaps (validation MAE) on the FACENAME task-fMRI. Each panel fixes $K_{\mathrm{REC}}$ and varies $K_{\mathrm{ENC}}$ and $K_{\mathrm{DIST}}$.}
\label{fig:k_sensitivity}
\vspace{-4mm}
\end{figure}

\begin{table}[t]
\caption{Distance metric sensitivity on the FACENAME task.}
\vspace{-1.5mm}
\centering
\renewcommand{\arraystretch}{1}
\setlength{\tabcolsep}{3.8pt}
\scriptsize
\resizebox{0.75\linewidth}{!}{%
\begin{tabular}{l c c}
\toprule
\textbf{Distance} 
& MAE 
& $R^2$ \\
\midrule

\rowcolor{gray!15}
\rule{0pt}{2ex}%
Euclidean
& $\mathbf{7.83\!\pm\!0.10}$ 
& $\mathbf{0.439\!\pm\!0.026}$ \\

Cosine
& $7.94\!\pm\!0.08$
& $0.404\!\pm\!0.023$ \\

Correlation
& $7.96\!\pm\!0.09$
& $0.414\!\pm\!0.021$ \\

\bottomrule
\end{tabular}%
}
\label{tab:distance_sensitivity}
\vspace{-4mm}
\end{table}

\begin{figure*}[t]
    \centering
    \includegraphics[page=1, width=0.9\textwidth]{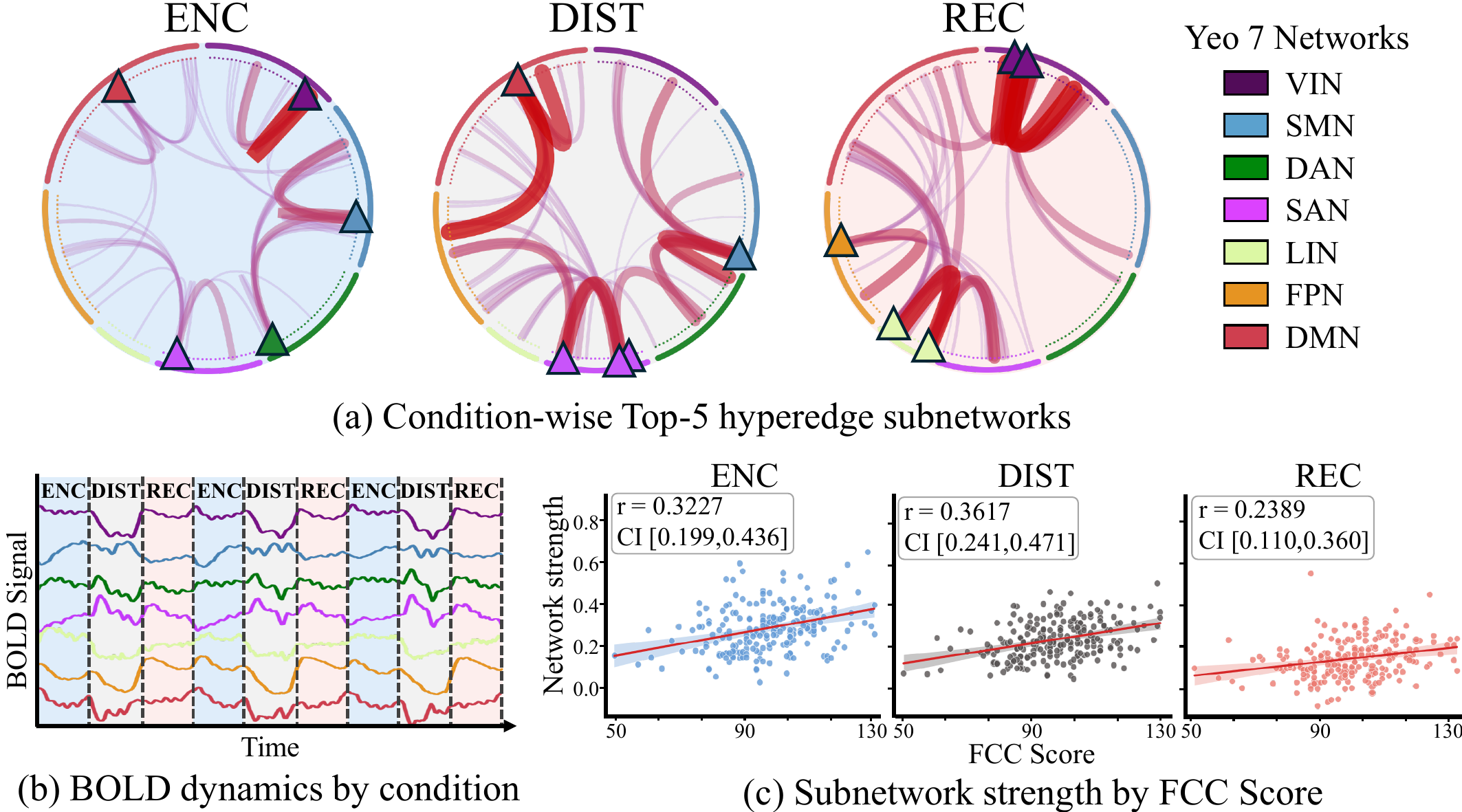}
    \vspace{-2mm}
    \caption{Interpretability analyses at the Yeo-7 networks: Visual (\textcolor{yeoVis}{\textbf{VIN}}), Somatomotor (\textcolor{yeoSomMot}{\textbf{SMN}}), Dorsal Attention (\textcolor{yeoDorsAttn}{\textbf{DAN}}), Salience/Ventral Attention (\textcolor{yeoSalVentAttn}{\textbf{SAN}}), Limbic (\textcolor{yeoLimbic}{\textbf{LIN}}),  FrontoParietal (\textcolor{yeoCont}{\textbf{FPN}}), and Default Mode (\textcolor{yeoDefault}{\textbf{DMN}}).
    (a) Condition-wise top-5 hyperedge subnetworks; triangles mark anchors, and ribbons summarize dominant network pairs.
    (b) Condition-wise BOLD signal.
    (c) Subnetwork strength by FCC for each condition.}
    \label{fig:interpretability}
    \vspace{-4mm}
\end{figure*}

Here, $K(\mathrm{ENC},\mathrm{DIST},\mathrm{REC})=(K_{\mathrm{ENC}},K_{\mathrm{DIST}},K_{\mathrm{REC}})$; Top-3 settings are selected by fold-wise validation MAE and reported as test mean$\pm$std across folds. All hyperparameter and checkpoint selections, including \(K_q\), were performed exclusively on the inner validation split; the outer test fold was held out for final evaluation only. Table~\ref{tab:facename_k_settings} compares a shared-$K$ design with the proposed condition-specific $K_q$ design. Under shared-$K$, the best configuration is $(5,5,5)$, achieving $7.98\!\pm\!0.09$ MAE and $0.408\!\pm\!0.036$ $R^2$. In contrast, condition-wise $K_q$ attains its best result at $(50,45,5)$ with $7.83\!\pm\!0.10$ MAE and $0.439\!\pm\!0.026$ $R^2$. Moreover, all top-3 condition-wise configurations outperform the top shared-$K$ settings, suggesting that the gain is not driven by a single isolated configuration.

Fig.~\ref{fig:k_sensitivity} further visualizes the $K_q$ sensitivity landscape by fixing $K_{\mathrm{REC}}\in\{5,30,50\}$ and varying $K_{\mathrm{ENC}}$ and $K_{\mathrm{DIST}}$. The lowest-error regions are concentrated when $K_{\mathrm{REC}}$ is small, while high-performing configurations for ENC and DIST generally appear in large neighborhood ranges. Although the performance surface is not strictly monotonic, the asymmetric top-performing configurations indicate that a single shared $K$ cannot fully capture the heterogeneous neighborhood scales across task phases.

\subsubsection{Distance metric sensitivity}


We compared Euclidean, cosine, and correlation similarities, changing only this metric while keeping the selected $K_q$ (Table~\ref{tab:facename_k_settings}), data splits, and all other settings fixed. As shown in Table~\ref{tab:distance_sensitivity}, Euclidean distance achieved the lowest MAE and highest $R^2$ among the three metrics, while cosine and correlation similarity degraded performance. These results support the use of negative squared Euclidean distance for constructing condition-wise FC-profile neighborhoods in CoHyFuse.



\subsubsection{Condition-wise FC reliability and sensitivity}
As an additional FC-estimator sensitivity check, we repeated the FACENAME experiment using Ledoit--Wolf shrinkage-regularized condition-wise FCs, while keeping the selected \(K_q\), data splits, and training settings unchanged. Performance remained close to the Pearson-FC setting (MAE \(=7.92\pm0.11\), \(R^2=0.425\pm0.025\)), suggesting that CoHyFuse is not solely dependent on the unregularized Pearson FC estimator. A within-subject split-half check (odd vs.\ even blocks) showed moderate-to-good edge-wise consistency of the condition-wise FCs (mean r=0.61), which exceeded the mean between-condition similarity (r=0.42), indicating that the state-resolved profiles are reliable yet condition-distinct rather than noise-dominated.


\subsubsection{Confound control}
\label{sec:confound}
Because FC is entangled with age and head motion \cite{power2012spurious}, we tested whether predictions reduce to measured demographic and motion covariates. The confound set was age/sex/FD for FACENAME and sex/FD for VISMOTOR (age is excluded as it is the VISMOTOR target). A confound-only model yielded $R^2 = 0.25$ (FACENAME) and $R^2 \approx 0$ (VISMOTOR), both below CoHyFuse. For the FACENAME confound-adjusted analysis, nuisance-regression coefficients for residualizing the FC-derived features and target were estimated within each outer fold using only the training split and then applied unchanged to validation/test subjects \cite{scheinost2019ten}. After this fold-wise residualization, the confound-adjusted association remained significant for FACENAME (partial $r = 0.34$, $p < .001$, 10{,}000 permutations), indicating that FACENAME prediction was not fully explained by measured age, sex, and motion covariates. For VISMOTOR, where age is itself the target, we instead verified that prediction error was not appreciably correlated with motion ($r = 0.09$), suggesting that age prediction was not strongly motion-driven.

\subsection{Interpretability Analysis}

To provide neurocognitively meaningful insights beyond predictive accuracy, we analyzed both condition-wise BOLD activity patterns and the hyperedges that were most influential for prediction in the FACENAME task-fMRI of the AABC dataset, with all summaries reported at the Yeo-7 network level \cite{yeo2011organization}.

\subsubsection{BOLD signal analysis}
The mean BOLD signal across subjects at the Yeo-7 network level exhibited condition-dependent differences (Fig.~\ref{fig:interpretability}(b)). The VIN showed high activity during ENC, reached its lowest level during DIST, and recovered during REC. In contrast, the SAN was relatively weak in ENC but became engaged in DIST. These large-scale shifts across task conditions suggest that the data contain distinct condition-dependent regimes. Therefore, treating all TRs as a single homogeneous sequence may mix conditions with different baseline activation contexts and blur condition-wise connectivity patterns\cite{greene2018task}.

\subsubsection{Condition-wise hyperedge occlusion analysis}
To interpret which condition-wise hyperedges were most influential for model prediction (Fig.~\ref{fig:interpretability}(a)), we ranked hyperedges via occlusion by removing one constructed hyperedge at a time and measuring the resulting increase in prediction error. Rankings were derived on the training split and assessed on a held-out test split, and hyperedge memberships were summarized using \(H_{\mathrm{mean}}\) aggregated to Yeo-7 networks.

\noindent\textbf{ENC (Encoding):}
In ENC, the Top-5 hyperedges were anchored evenly across five large-scale systems (VIN, DAN, SMN, SAN, and DMN; $n{=}1$ each). Dominant network-pair contributions were within-network VIN--VIN and SMN--SMN, with an additional cross-network SMN--DAN component. Neurocognitively, this pattern suggests that the model's predictions were most sensitive to within-VIN structure and SMN--DAN coupling, consistent with visual processing and goal-directed attention/response-selection demands during encoding \cite{sperling2001encoding}.

\noindent\textbf{DIST (Distraction):}
In DIST, anchors were concentrated in SAN ($n{=}3$; DMN $n{=}1$, SMN $n{=}1$), and network-pair contributions were SAN--FPN
and SAN--SAN, consistent with accounts in which the SAN is engaged by distracting events and interacts with FPN during distraction \cite{seeley2007dissociable}. Consistent with this motif, DIST subnetwork strength showed the strongest condition-wise association with FCC (Fig.~\ref{fig:interpretability}(c)).

\noindent\textbf{REC (Recall):}
In REC, anchors were distributed across VIN ($n{=}2$), LIN ($n{=}2$), and FPN ($n{=}1$).
Network-pair summaries highlighted
LIN--DMN and LIN--LIN
contributions alongside within-network VIN--VIN.
This configuration is compatible with episodic retrieval involving interactions between LIN cortical parcels and DMN components, alongside within-VIN structure and possible involvement of FPN in goal-directed recall \cite{spreng2010default}. Across task conditions, the occlusion-based anchors and the \(H_{\mathrm{mean}}\) network-pair summaries showed distinct, condition-wise patterns. These results reflect model reliance for prediction rather than causal necessity of the implicated networks.

\subsubsection{Validation of subnetwork strength}
To test whether condition-wise subnetwork strength captures individual differences in FCC, we examined its continuous association with FCC in the held-out test split (Fig.~\ref{fig:interpretability}(c)). Across conditions, strength was positively associated with FCC (ENC/DIST/REC: $r$=0.24--0.36), with the strongest association observed in DIST. 95\% confidence intervals were estimated via Fisher's $z$ transform, and statistical significance was assessed using permutation tests ($N_{\mathrm{p}}$=10,000) with false discovery rate (FDR) correction across conditions (all $q_{FDR}\le$ 0.018). These associations persisted after covariate adjustment (age, mean FD \cite{power2012spurious}).

\begin{table}[t]
\caption{Performance comparison on the CMI-HBN benchmark.}
\vspace{-1.5mm}
\centering
\renewcommand{\arraystretch}{1}
\setlength{\tabcolsep}{1pt}
\footnotesize
\resizebox{\linewidth}{!}{%
\begin{tabular}{lccccc}
\toprule
\textbf{Model}
& \makecell{mAUC (\%)} 
& \makecell{ACC (\%)} 
& \makecell{SEN$_N$ (\%)} 
& \makecell{SEN$_I$ (\%)} 
& \makecell{SEN$_C$ (\%)} \\
\midrule

SVM          
& 62.8{\scriptsize$\pm$2.5} 
& 63.6{\scriptsize$\pm$3.3} 
& 58.2{\scriptsize$\pm$5.1} 
& 62.1{\scriptsize$\pm$4.9} 
& 69.5{\scriptsize$\pm$5.3} \\

MLP         
& 63.9{\scriptsize$\pm$3.0} 
& 65.2{\scriptsize$\pm$3.8} 
& 61.2{\scriptsize$\pm$5.4} 
& 63.4{\scriptsize$\pm$5.1} 
& 70.2{\scriptsize$\pm$5.4} \\

GCN \cite{kipf2016semi}          
& 60.6{\scriptsize$\pm$3.4} 
& 61.2{\scriptsize$\pm$4.2} 
& 57.5{\scriptsize$\pm$5.8} 
& 59.8{\scriptsize$\pm$5.6} 
& 65.5{\scriptsize$\pm$6.0} \\

GAT \cite{velivckovic2017graph}          
& 61.2{\scriptsize$\pm$3.2} 
& 62.6{\scriptsize$\pm$4.1} 
& 59.3{\scriptsize$\pm$5.5} 
& 60.8{\scriptsize$\pm$5.4} 
& 66.9{\scriptsize$\pm$5.8} \\

HGNN \cite{feng2019hypergraph}         
& 67.1{\scriptsize$\pm$2.8} 
& 68.4{\scriptsize$\pm$3.5} 
& 65.3{\scriptsize$\pm$4.7} 
& 66.7{\scriptsize$\pm$4.6} 
& 72.6{\scriptsize$\pm$5.0} \\

HNHN \cite{dong2020hnhn}
& 66.8{\scriptsize$\pm$2.8} 
& 68.0{\scriptsize$\pm$3.6} 
& 65.1{\scriptsize$\pm$4.8} 
& 66.4{\scriptsize$\pm$4.6} 
& 71.8{\scriptsize$\pm$5.1} \\

BrainGNN \cite{li2021braingnn}     
& 66.0{\scriptsize$\pm$3.4} 
& 67.4{\scriptsize$\pm$4.4} 
& 64.4{\scriptsize$\pm$5.9} 
& 66.0{\scriptsize$\pm$5.7} 
& 71.1{\scriptsize$\pm$6.2} \\

ED-HNN \cite{wang2022equivariant}  
& 66.6{\scriptsize$\pm$2.9} 
& 68.7{\scriptsize$\pm$3.7} 
& 65.8{\scriptsize$\pm$5.0} 
& 67.3{\scriptsize$\pm$4.7} 
& 72.5{\scriptsize$\pm$5.2} \\

ALTER \cite{yu2024long} 
& 67.0{\scriptsize$\pm$2.6} 
& \underline{71.6{\scriptsize$\pm$3.3}} 
& 66.8{\scriptsize$\pm$4.5} 
& \underline{69.2{\scriptsize$\pm$4.3}} 
& \underline{77.7{\scriptsize$\pm$4.7}} \\

TA-GAT \cite{wang2025towards}
& 67.8{\scriptsize$\pm$2.5} 
& 70.5{\scriptsize$\pm$3.2} 
& 67.5{\scriptsize$\pm$4.3} 
& 68.3{\scriptsize$\pm$4.1} 
& 75.1{\scriptsize$\pm$4.6} \\

STNAGNN \cite{wang2026stnagnn}
& \underline{68.0{\scriptsize$\pm$2.4}} 
& 71.1{\scriptsize$\pm$3.1} 
& \underline{68.1{\scriptsize$\pm$4.2}} 
& 69.0{\scriptsize$\pm$4.0} 
& 75.5{\scriptsize$\pm$4.4} \\

\rowcolor{gray!15}
\textbf{CoHyFuse}
\rule{0pt}{2ex}
& \textbf{72.0{\scriptsize$\pm$2.1}} 
& \textbf{74.2{\scriptsize$\pm$2.9}} 
& \textbf{71.3{\scriptsize$\pm$3.8}} 
& \textbf{72.1{\scriptsize$\pm$3.7}} 
& \textbf{78.4{\scriptsize$\pm$4.1}} \\

\bottomrule

\end{tabular}%
}
\label{tab:clinical_cmi_hbn}
\vspace{-4mm}

\end{table}

\subsection{Application to CMI-HBN Dataset}

To assess the generality of CoHyFuse beyond the AABC regression benchmarks, we further applied it to an exploratory three-class ADHD classification benchmark in the CMI-HBN pediatric cohort. The task distinguishes NoDx, ADHD-I, and ADHD-C using REST and PEER-state fMRI. For this classification benchmark, the scalar regression head was replaced with a three-way softmax classifier trained using cross-entropy loss, while the condition-wise hypergraph construction and fusion modules were kept unchanged. All methods were trained and evaluated using identical subject-wise splits and the same ROI time-series inputs. Here, mAUC and ACC denote macro-AUC and accuracy, respectively, and SEN$_N$/SEN$_I$/SEN$_C$ denote sensitivities for NoDx/ADHD-I/ADHD-C.

As shown in Table~\ref{tab:clinical_cmi_hbn}, CoHyFuse achieved the best overall performance, with 72.0\% mAUC and 74.2\% accuracy. Compared with the strongest baseline, CoHyFuse improved mAUC by 4.0 percentage points over STNAGNN and accuracy by 2.6 percentage points over ALTER. CoHyFuse achieved the highest sensitivity across all three diagnostic groups, suggesting that condition-wise hypergraph-based ROI-set fusion provides useful predictive information for pediatric ADHD subtype classification.

\vspace{-4pt}
\section{Discussion}

CoHyFuse was designed to preserve task-state information that is often lost when task-fMRI is summarized as a single condition-agnostic FC graph. Across AABC FACENAME and VISMOTOR benchmarks, CoHyFuse achieved the best mean performance among the evaluated baselines, and ablations showed that both the condition-wise hypergraph branch and the whole-session FC branch contributed to prediction. These results suggest that task-state reconfiguration and session-level connectivity provide complementary information, consistent with prior work showing that task-fMRI can amplify behaviorally relevant individual differences and that task-state FC carries functional information beyond resting-state connectivity \cite{greene2018task}. The operator-control and \(K_q\) analyses further support modeling task-fMRI beyond pairwise ROI--ROI graphs: replacing HGNN with GCN/GAT variants reduced performance, and condition-specific hypergraph construction was more effective than a shared neighborhood scale. This is consistent with prior hypergraph connectomics evidence that multi-ROI topological patterns can carry task- and behavior-relevant information \cite{yuan2024mhsa}. In CoHyFuse, the condition-wise hypergraph branch captures state-dependent FC-profile neighborhood structure, whereas the whole-session FC branch provides a compact summary of session-level connectivity.

Hyperedge occlusion and subnetwork-strength analyses highlighted SAN--FPN and within-SAN motifs, and DIST subnetwork strength showed the strongest association with FCC after age and motion adjustment. These findings suggest a model-relevant salience--control pattern during distraction, but should be interpreted as model reliance rather than causal biomarker evidence. The CMI-HBN ADHD experiment provides an auxiliary clinical-task evaluation, suggesting that the same state-wise fusion mechanism can be applied beyond adult cognitive regression under matched splits and ROI inputs. Limitations include reliance on paradigms with sufficiently long condition windows for stable condition-wise FC estimation and increased hyperparameter search cost for condition-wise \(K_q\). Future work will explore learnable neighborhood selection, temporally adaptive condition modeling, and broader validation across tasks, cohorts, atlases, and connectivity estimators.


\vspace{-8pt}
\section{Conclusion}
We presented CoHyFuse, a condition-aware ROI-centered hypergraph fusion framework for task-fMRI prediction. CoHyFuse constructs condition-wise top-\(K_q\) hypergraphs and fuses HGNN-based condition embeddings with a complementary whole-session FC branch. Experiments on AABC and CMI-HBN benchmarks showed the best mean performance among evaluated baselines, and ablations supported the contributions of condition-specific hypergraph construction, heterogeneous \(K_q\), and dual-branch fusion. Interpretability analysis further identified DIST-related SAN--FPN and SAN--SAN motifs as model-relevant patterns associated with FCC. Overall, CoHyFuse provides a compact and interpretable framework for modeling state-resolved hypergraph-defined ROI-set structure in task-fMRI.

\let\oldthebibliography\thebibliography
\renewcommand{\thebibliography}[1]{%
  \oldthebibliography{#1}%
  \fontsize{8.0pt}{8.4pt}\selectfont
  \setlength{\itemsep}{-1.2pt}%
  \setlength{\parsep}{0pt}%
  \setlength{\topsep}{0pt}%
}
\vspace{-8pt}
\bibliographystyle{unsrt}

\bibliography{ref}

@article{Bookheimer2019_HCPAging,
  title   = {The Lifespan Human Connectome Project in Aging: An overview},
  author  = {Bookheimer, Susan Y. and Salat, David H. and Terpstra, Melissa and Ances, Beau M. and Barch, Deanna M. and Buckner, Randy L. and others},
  journal = {NeuroImage},
  volume  = {185},
  pages   = {335--348},
  year    = {2019},
  doi     = {10.1016/j.neuroimage.2018.10.009}
}

@article{schaefer2018local,
  title={Local-global parcellation of the human cerebral cortex from intrinsic functional connectivity {MRI}},
  author={Schaefer, Alexander and Kong, Ru and Gordon, Evan M and Laumann, Timothy O and Zuo, Xi-Nian and Holmes, Avram J and Eickhoff, Simon B and Yeo, BT Thomas},
  journal={Cerebral cortex},
  volume={28},
  number={9},
  pages={3095--3114},
  year={2018},
  publisher={Oxford University Press}
}

@article{akshoomoff2013viii,
  title={VIII. {NIH} Toolbox Cognition Battery (CB): composite scores of crystallized, fluid, and overall cognition},
  author={Akshoomoff, Natacha and Beaumont, Jennifer L and Bauer, Patricia J and Dikmen, Sureyya S and Gershon, Richard C and Mungas, Dan and Slotkin, Jerry and Tulsky, David and Weintraub, Sandra and Zelazo, Philip David and others},
  journal={Monographs of the Society for Research in Child Development},
  volume={78},
  number={4},
  pages={119--132},
  year={2013},
  publisher={Wiley Online Library}
}

@article{kipf2016semi,
  title={Semi-supervised classification with graph convolutional networks},
  author={Kipf, Thomas N and Welling, Max},
  journal={arXiv preprint arXiv:1609.02907},
  year={2016}
}

@article{velivckovic2017graph,
  title={Graph attention networks},
  author={Velickovic, Petar and Cucurull, Guillem and Casanova, Arantxa and Romero, Adriana and Lio, Pietro and Bengio, Yoshua},
  journal={arXiv preprint arXiv:1710.10903},
  year={2017}
}

@article{li2021braingnn,
  title={{BrainGNN}: Interpretable brain graph neural network for {fMRI} analysis},
  author={Li, Xiaoxiao and Zhou, Yuan and Dvornek, Nicha and Zhang, Muhan and Gao, Siyuan and Zhuang, Juntang and Scheinost, Dustin and Staib, Lawrence H and Ventola, Pamela and Duncan, James S},
  journal={Medical Image Analysis},
  volume={74},
  pages={102233},
  year={2021},
  publisher={Elsevier}
}

@article{yu2024long,
  title={Long-range brain graph transformer},
  author={Yu, Shuo and Jin, Shan and Li, Ming and Sarwar, Tabinda and Xia, Feng},
  journal={Advances in Neural Information Processing Systems},
  volume={37},
  pages={24472--24495},
  year={2024}
}

@inproceedings{feng2019hypergraph,
  title={Hypergraph neural networks},
  author={Feng, Yifan and You, Haoxuan and Zhang, Zizhao and Ji, Rongrong and Gao, Yue},
  booktitle={Proceedings of the AAAI conference on artificial intelligence},
  volume={33},
  number={01},
  pages={3558--3565},
  year={2019}
}

@article{dong2020hnhn,
  title={Hnhn: Hypergraph networks with hyperedge neurons},
  author={Dong, Yihe and Sawin, Will and Bengio, Yoshua},
  journal={arXiv preprint arXiv:2006.12278},
  year={2020}
}

@article{wang2022equivariant,
  title={Equivariant hypergraph diffusion neural operators},
  author={Wang, Peihao and Yang, Shenghao and Liu, Yunyu and Wang, Zhangyang and Li, Pan},
  journal={arXiv preprint arXiv:2207.06680},
  year={2022}
}

@article{greene2018task,
  title={Task-induced brain state manipulation improves prediction of individual traits},
  author={Greene, Abigail S and Gao, Siyuan and Scheinost, Dustin and Constable, R Todd},
  journal={Nature communications},
  volume={9},
  number={1},
  pages={2807},
  year={2018},
  publisher={Nature Publishing Group UK London}
}

@article{sperling2001encoding,
  title={Encoding novel face-name associations: A functional {MRI} study},
  author={Sperling, Reisa A and Bates, Julianna F and Cocchiarella, Andrew J and Schacter, Daniel L and Rosen, Bruce R and Albert, Marilyn S},
  journal={Human brain mapping},
  volume={14},
  number={3},
  pages={129--139},
  year={2001},
  publisher={Wiley Online Library}
}

@article{seeley2007dissociable,
  title={Dissociable intrinsic connectivity networks for salience processing and executive control},
  author={Seeley, William W and Menon, Vinod and Schatzberg, Alan F and Keller, Jennifer and Glover, Gary H and Kenna, Heather and Reiss, Allan L and Greicius, Michael D},
  journal={Journal of neuroscience},
  volume={27},
  number={9},
  pages={2349--2356},
  year={2007},
  publisher={Society for Neuroscience}
}

@article{spreng2010default,
  title={Default network activity, coupled with the frontoparietal control network, supports goal-directed cognition},
  author={Spreng, R Nathan and Stevens, W Dale and Chamberlain, Jon P and Gilmore, Adrian W and Schacter, Daniel L},
  journal={Neuroimage},
  volume={53},
  number={1},
  pages={303--317},
  year={2010},
  publisher={Elsevier}
}

@article{yeo2011organization,
  title={The organization of the human cerebral cortex estimated by intrinsic functional connectivity},
  author={Yeo, BT Thomas and Krienen, Fenna M and Sepulcre, Jorge and Sabuncu, Mert R and Lashkari, Danial and Hollinshead, Marisa and Roffman, Joshua L and Smoller, Jordan W and Z{\"o}llei, Lilla and Polimeni, Jonathan R and others},
  journal={Journal of neurophysiology},
  year={2011},
  publisher={American Physiological Society Bethesda, MD}
}

@article{vaswani2017attention,
  title={Attention is all you need},
  author={Vaswani, Ashish and Shazeer, Noam and Parmar, Niki and Uszkoreit, Jakob and Jones, Llion and Gomez, Aidan N and Kaiser, {\L}ukasz and Polosukhin, Illia},
  journal={Advances in neural information processing systems},
  volume={30},
  year={2017}
}

@article{power2012spurious,
  title={Spurious but systematic correlations in functional connectivity {MRI} networks arise from subject motion},
  author={Power, Jonathan D and Barnes, Kelly A and Snyder, Abraham Z and Schlaggar, Bradley L and Petersen, Steven E},
  journal={Neuroimage},
  volume={59},
  number={3},
  pages={2142--2154},
  year={2012},
  publisher={Elsevier}
}

@article{glasser2013minimal,
  title={The minimal preprocessing pipelines for the Human Connectome Project},
  author={Glasser, Matthew F and Sotiropoulos, Stamatios N and Wilson, J Anthony and Coalson, Timothy S and Fischl, Bruce and Andersson, Jesper L and Xu, Junqian and Jbabdi, Saad and Webster, Matthew and Polimeni, Jonathan R and others},
  journal={Neuroimage},
  volume={80},
  pages={105--124},
  year={2013},
  publisher={Elsevier}
}

@inproceedings{yuan2024mhsa,
  title={MHSA: a multi-scale hypergraph network for mild cognitive impairment detection via synchronous and attentive fusion},
  author={Yuan, Manman and Jia, Weiming and Luo, Xiong and Ye, Jiazhen and Zhu, Peican and Li, Junlin},
  booktitle={2024 IEEE International Conference on Bioinformatics and Biomedicine (BIBM)},
  pages={2808--2815},
  year={2024},
  organization={IEEE}
}

@article{alexander2017open,
  title={An open resource for transdiagnostic research in pediatric mental health and learning disorders},
  author={Alexander, Lindsay M and Escalera, Jasmine and Ai, Lei and Andreotti, Charissa and Febre, Karina and Mangone, Alexander and Vega-Potler, Natan and Langer, Nicolas and Alexander, Alexis and Kovacs, Meagan and others},
  journal={Scientific data},
  volume={4},
  number={1},
  pages={170181},
  year={2017},
  publisher={Nature Publishing Group}
}

@article{wang2026stnagnn,
  title={{STNAGNN}: Data-driven spatio-temporal brain connectivity beyond {FC}},
  author={Wang, Jiyao and Dvornek, Nicha C and Duan, Peiyu and Staib, Lawrence H and Ventola, Pamela and Duncan, James S},
  journal={Proceedings of machine learning research},
  volume={301},
  pages={1688},
  year={2026}
}

@inproceedings{wang2025towards,
  title={Towards Zero-Shot Task-Generalizable Learning on {fMRI}},
  author={Wang, Jiyao and Dvornek, Nicha C and Duan, Peiyu and Staib, Lawrence H and Duncan, James S},
  booktitle={2025 IEEE 22nd International Symposium on Biomedical Imaging (ISBI)},
  pages={1--5},
  year={2025},
  organization={IEEE}
}

@article{scheinost2019ten,
  title={Ten simple rules for predictive modeling of individual differences in neuroimaging},
  author={Scheinost, Dustin and Noble, Stephanie and Horien, Corey and Greene, Abigail S and Lake, Evelyn MR and Salehi, Mehraveh and Gao, Siyuan and Shen, Xilin and O'Connor, David and Barron, Daniel S and others},
  journal={NeuroImage},
  volume={193},
  pages={35--45},
  year={2019},
  publisher={Elsevier}
}

@inproceedings{wang2023adaptive,
  title={Adaptive particle swarm architecture search based on multi-level convolutions for functional brain network classification},
  author={Wang, Xingyu and Ji, Junzhong},
  booktitle={2023 IEEE International Conference on Bioinformatics and Biomedicine (BIBM)},
  pages={654--659},
  year={2023},
  organization={IEEE}
}

\end{document}